\documentclass[11pt,a4paper]{article}
\usepackage[hyperref]{emnlp2020}
\usepackage{times}
\usepackage{latexsym}

\usepackage{microtype}

\usepackage{booktabs} 

\usepackage{graphicx}

\aclfinalcopy 
\def\aclpaperid{1165} 

\makeatletter
\def\@fnsymbol#1{\ensuremath{\ifcase#1\or \dag\or \ddagger\or
   \mathsection\or \mathparagraph\or \|\or **\or \dagger\dagger
   \or \ddagger\ddagger \else\@ctrerr\fi}}
    \makeatother

\title{Investigating Transferability in Pretrained Language Models}

\author{
  Alex Tamkin\thanks{\texttt{ atamkin@stanford.edu}}\\
  Stanford University \\\And
  Trisha Singh\\
  Stanford University  \\\And
  Davide Giovanardi\\
  Stanford University \\\And
  Noah Goodman\\
  Stanford University
  }

\date{}

\begin{document}

\maketitle
\begin{abstract}
  \label{abstract}

How does language model pretraining help transfer learning? 
We consider a simple ablation technique for determining the impact of each pretrained layer on transfer task performance. 
This method, \emph{partial reinitialization}, involves replacing different layers of a pretrained model with random weights, then finetuning the entire model on the transfer task and observing the change in performance. 
This technique reveals that in BERT, layers with high probing performance on downstream GLUE tasks are \emph{neither necessary nor sufficient} for high accuracy on those tasks.
Furthermore, the benefit of using pretrained parameters for a layer varies dramatically with finetuning dataset size: parameters that provide tremendous performance improvement when data is plentiful may provide negligible benefits in data-scarce settings.
These results reveal the complexity of the transfer learning process, highlighting the limitations of methods that operate on frozen models or single data samples.
\end{abstract}

\section{Introduction}
\label{sec:intro}

Despite the striking success of transfer learning in NLP, remarkably little is understood about how these pretrained models improve downstream task performance. Recent work on understanding deep NLP models has centered on \emph{probing}, a methodology that involves training classifiers for different tasks on model representations \citep{alain2016understanding, conneau2018you, hupkes2018visualisation, liu2019linguistic, tenney2019bert, tenney2019you, goldberg2019assessing, hewitt2019structural}. 
While probing aims to uncover what a network has already learned, a major goal of machine learning is \emph{transfer}: systems that build upon what they have learned to expand what they \emph{can} learn. Given that most recent models are updated end-to-end during finetuning \citep[e.g.][]{devlin2018bert, howard2018universal, radford2018improving}, 
it is unclear how, or even whether, the knowledge uncovered by probing contributes to these models' transfer learning success.  

\begin{figure}
    \includegraphics[width=\columnwidth]{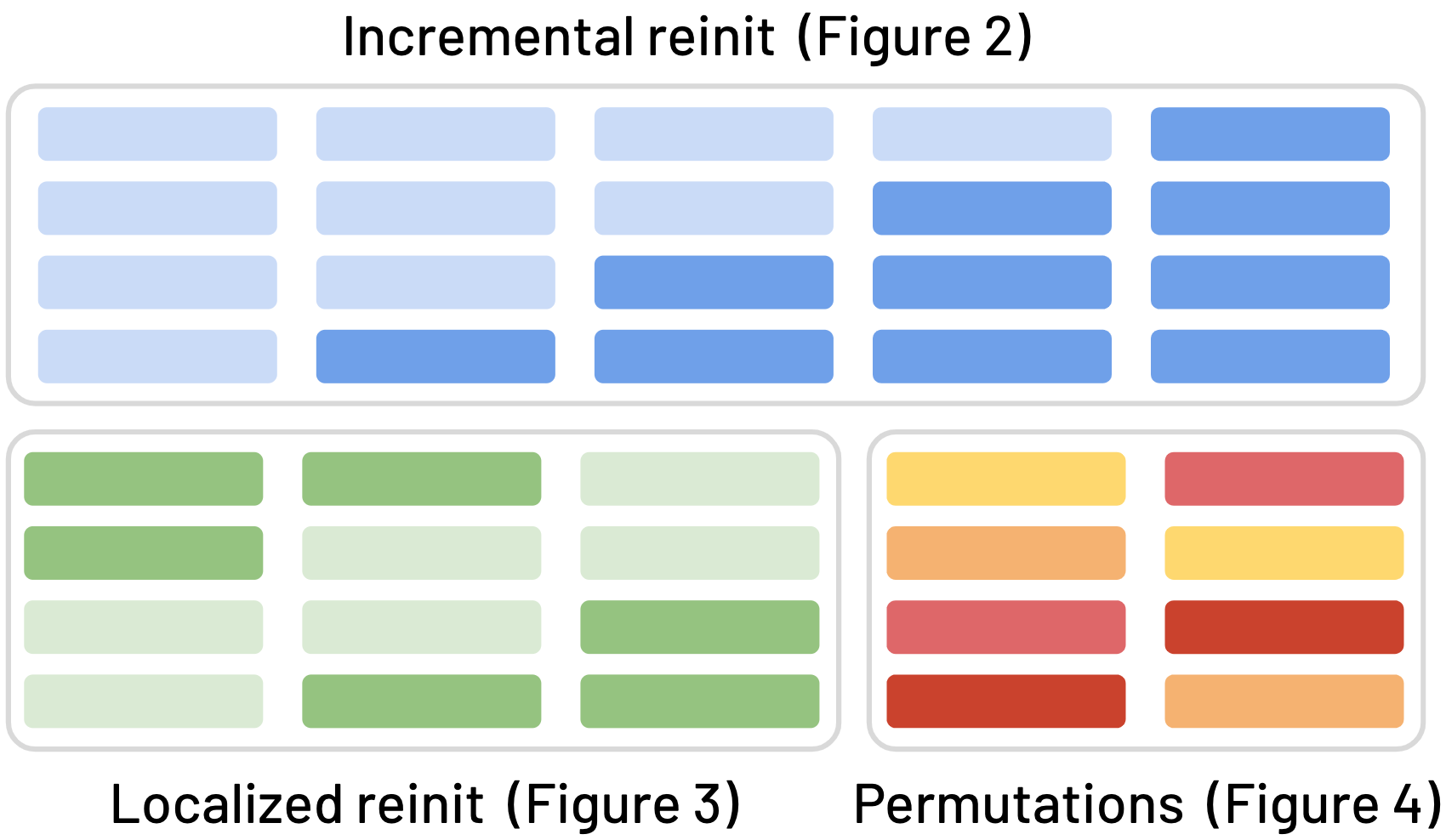}
    \caption{\textbf{The three experiments we explore.} Lighter shades indicate randomly reinitialized layers, while darker shades indicate layers with BERT parameters. For layer permutations, all layers hold BERT parameters, what changes between trials is their order. In all three experiments, the entire model is finetuned end-to-end on the GLUE task.}
    \label{fig:rects}
\end{figure}

\begin{figure}
    \includegraphics[width=\columnwidth]{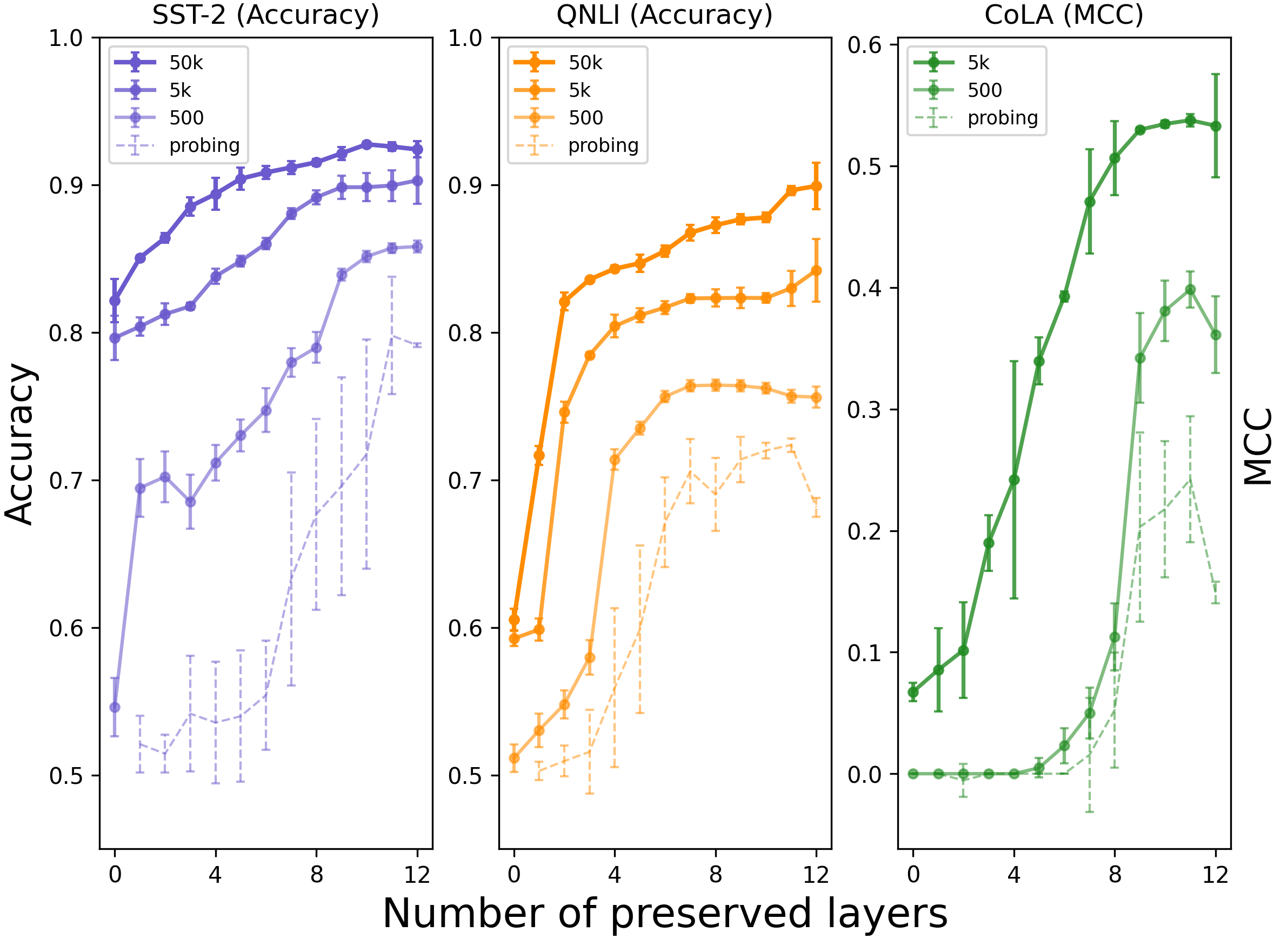}
    \caption{\textbf{The benefit of using BERT parameters instead of random parameters at a particular layer varies dramatically depending on the size of the finetuning dataset. However, as finetuning dataset size decreases, the curves align more closely with probing performance at each layer.} Solid lines show finetuning results after reinitializing all layers past layer $k$ in BERT-Base. 12 shows the full BERT model, while 0 shows a model with all layers reinitialized. Line darkness indicates subsampled dataset size. The dashed lines show probing performance at each layer. Error bars are 95\% CIs.}
    \label{fig:progressive}
\end{figure}

In a sense, probing can be seen as quantifying the \emph{transferability of representations} from one task to another, as it measures how well a simple model (e.g., a softmax classifier) can perform the second task using only features from a model trained on the first. However, when pretrained models are finetuned end-to-end on a downstream task, what is transferred is not the features from each layer of the pretrained model, but its \emph{parameters}, which define a sequence of functions for processing representations. Critically, these functions and their interactions may shift considerably during training, potentially enabling higher performance despite not initially extracting features correlated with this task. We refer to this phenomenon of how layer parameters from one task can help transfer learning on another task as \emph{transferability of parameters}.

In this work, we investigate a methodology for measuring the transferability of different layer parameters in a pretrained language model to different transfer tasks, using BERT \citep{devlin2018bert} as our subject of analysis. Our methods, described more fully in Section \ref{sec:progressive} and Figure \ref{fig:rects}, involve \emph{partially reinitializing} BERT: replacing different layers with random weights and then observing the change in task performance after finetuning the entire model end-to-end. Compared to possible alternatives like freezing parts of the network or removing layers, partial reinitialization enables fairer comparisons by keeping the network's architecture and capacity constant between trials, changing only the parameters at initialization. Through experiments across different layers, tasks, and dataset sizes, this approach enables us to shed light on multiple dimensions of the transfer learning process: Are the early layers of the network more important than later ones for transfer learning? Do individual layers become more or less critical depending on the task or amount of finetuning data? Does the position of a particular layer within the network matter, or do its parameters aid optimization regardless of where they are in the network?

We find that when finetuning on a new task:
\begin{enumerate}
    \item Transferability of BERT layers varies dramatically depending on the amount of finetuning data available. Thus, claims that certain layers are universally responsible or important for learning certain linguistic tasks should be treated with caution. (Figure \ref{fig:progressive})
    \item Transferability of BERT layers is not in general predicted by the layer's probing performance for that task. However, as finetuning dataset size decreases, the two quantities exhibit a greater correspondence. (Figure \ref{fig:progressive}, dashed lines)
    \item Even holding dataset size constant, the most transferable BERT layers differ by task: for some tasks, only the early layers are important, while for others the benefits are more distributed across layers. (Figure \ref{fig:localized})
    \item Reordering the pretrained BERT layers before finetuning decreases downstream accuracy significantly, confirming that pretraining does not simply provide better-initialized individual layers; instead, transferability through learned interactions \emph{across layers} is crucial to the success of finetuning. (Figure \ref{fig:scrambles})
\end{enumerate}

\section{How many pretrained layers are necessary for finetuning?}
\label{sec:progressive}

Our first set of experiments aims to uncover how many pretrained layers are sufficient for accurate learning of a downstream task. To do this, we perform a series of \textbf{incremental reinitialization} experiments, where we reinitialize all layers after the $k$th layer of BERT-Base, for values $k \in \{0, 1, \ldots 12\}$, replacing them with random weights. We then finetune the entire model end-to-end on the target task. Note that $k=0$ corresponds to a BERT model with all layers reinitialized, while $k=12$ is the original BERT model. We do not reinitialize the BERT word embeddings. As BERT uses residual connections \citep{he2016deep} around layers, the model can simply learn to ignore any of the reinitialized layers if they are not helpful during finetuning.

We use the BERT-Base uncased model, implemented in PyTorch \citep{paszke2019pytorch} via the Transformers library \citep{wolf2019transformers}. We finetune the network using Adam \citep{kingma2014adam}, with a batch size of 8, a learning rate of \mbox{2e-5}, and default parameters otherwise. More details about reinitialization, training, statistical significance, and other methodological choices can be found in the Appendix. 
We conduct our experiments on three English language tasks from the GLUE benchmark, spanning the domains of sentiment, reasoning, and syntax \citep{wang2018glue}:

\paragraph{SST-2} Stanford Sentiment Treebank involves binary classification of a single sentence from a movie review as positive or negative \citep{socher2013recursive}.
\paragraph{QNLI} Question Natural Language Inference is a binary classification task derived from SQuAD \citep{rajpurkar2016squad, wang2018glue}. The task requires determining whether for a given (\textsc{question}, \textsc{answer}) pair the \textsc{question} is answered by the \textsc{answer}.
\paragraph{CoLA} The Corpus of Linguistic Acceptability is a binary classification task that requires determining whether a single sentence is linguistically acceptable \citep{warstadt2019neural}.

Because pretraining appears to be especially helpful in the small-data regime \citep{peters2018deep}, it is crucial to isolate task-specific effects from data quantity effects by controlling for finetuning dataset size. To do this, we perform our incremental reinitializations on randomly-sampled subsets of the data: 500, 5k, and 50k examples (excluding 50k for CoLA, which contains only 8.5k examples). The 5k subset size is then used as the default for our other experiments. To ensure that an unrepresentative sample is not chosen by chance, we run multiple trials with different subsamples. Confidence intervals produced through multiple trials also demonstrate that trends hold regardless of intrinsic task variability. 

While similar reinitialization schemes have been explored by \citet{yosinski2014transferable, raghu2019transfusion} in computer vision and briefly by \citet{radford2018improving} in an NLP context, none investigate these data quantity- and task-specific effects.

Figure \ref{fig:progressive} shows the results of our incremental reinitialization experiments. These results show that the transferability of a BERT layer varies dramatically based on the finetuning dataset size. Across all but the 500 example trials of SST-2, a more specific trend holds: earlier layers provide more of an improvement on finetuning performance when the finetuning dataset is large. This trend suggests that larger finetuning datasets may enable the network to learn a substitute for the parameters in the middle and later layers. In contrast, smaller datasets may leave the network reliant on existing feature processing in those layers. However, across all tasks and dataset sizes, it is clear that the pretrained parameters by themselves do not determine the impact they will have on finetuning performance: instead, a more complex interaction occurs between the parameters, optimizer, and the available data.

\section{Does probing predict layer transferability?}
\label{sec:probing}

What is the relationship between transferability of representations, measured by probing, and transferability of parameters, measured by partial reinitialization? To compare, we conduct probing experiments for our finetuning tasks on each layer of the pretrained BERT model. Our probing model averages each layer's hidden states, then passes the pooled representation through a linear layer and softmax to produce probabilities for each class. These task-specific components are identical to those in our reinitialization experiments; however, we keep the BERT model's parameters frozen when training our probes.

Our results, presented in Figure \ref{fig:progressive} (dashed lines), show a significant difference between the layers with the highest probing performance and reinitialization curves for the data-rich settings (darkest solid lines). For example, the probing accuracy on all tasks is near chance for the first six layers.
Despite this, these early layer parameters exhibit significant transferability to the finetuning tasks: preserving them while reinitializing all other layers enables large gains in finetuning accuracy across tasks. 
Interestingly, however, we observe that the smallest-data regime's curves are much more similar to the probing curves across all tasks than the larger-data regimes. Smaller finetuning datasets enable fewer updates to the network before overfitting occurs; thus, it may be that finetuning interpolates between the extremes of probing (no data) and fully-supervised learning (enough data to completely overwrite the pretrained parameters). We leave a more in-depth exploration of this connection to future work.

\section{Which layers are most useful for finetuning?}
\label{sec:localized}

\begin{figure}
    \includegraphics[width=1\columnwidth]{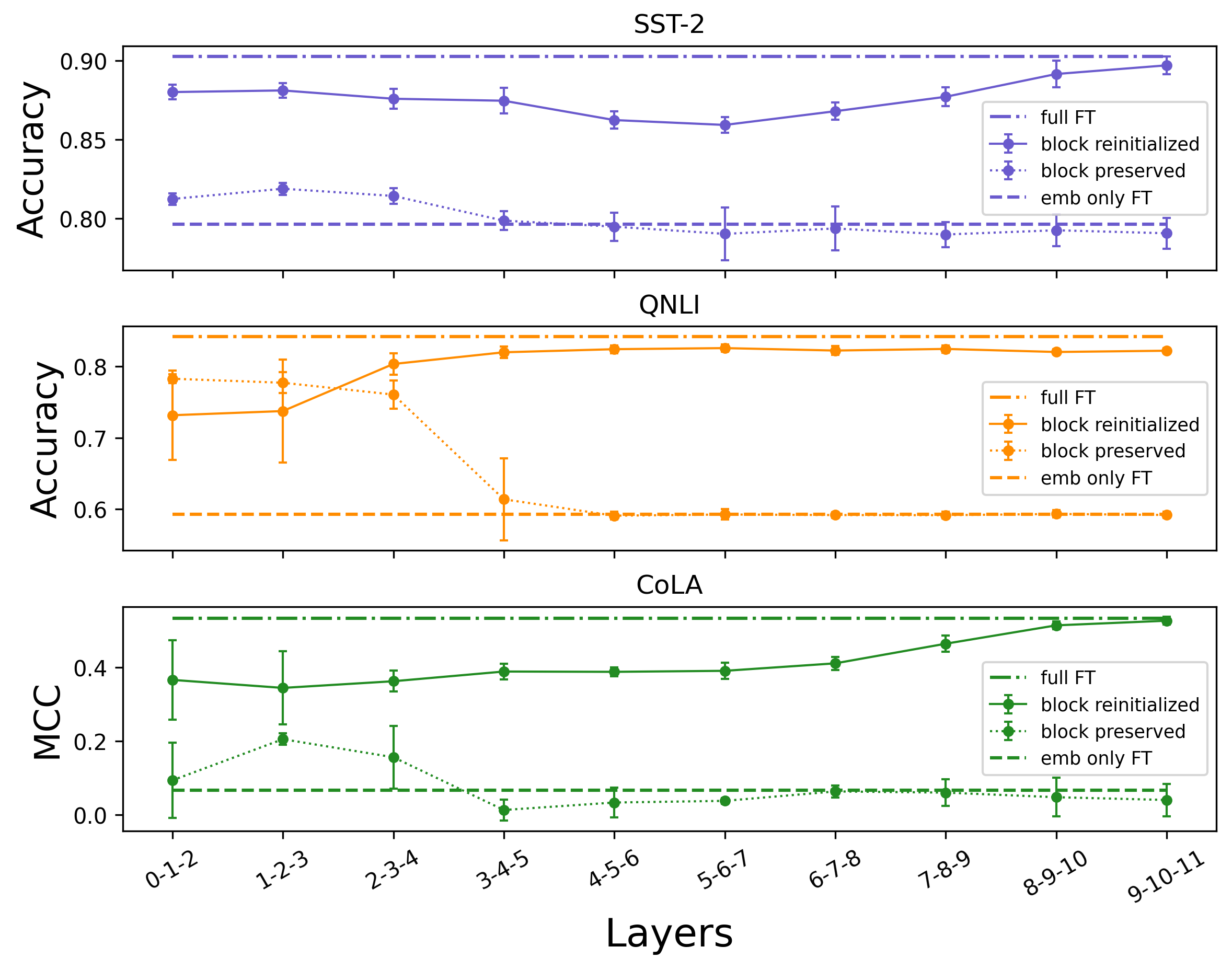}
    \caption{\textbf{Early layers provide the most QNLI gains, but middle ones yield an added boost for CoLA and SST-2.} Finetuning results for 1) reinitializing a consecutive three-layer block (``block reinitialized'') and 2) reinitializing all other layers (``block preserved''). Dashed horizontal lines show the finetuning performance of the full BERT model and the performance of a model with only embedding parameters preserved. Finetuning trials with 5k examples. Error bars are 95\% CIs.}
    \label{fig:localized}
\end{figure}

While the incremental reinitializations measure each BERT layer's incremental effect on transfer learning, they do not assess each layer's contribution in isolation, relative to either the full BERT model or an entirely reinitialized model. Measuring this requires eliminating the \emph{number} of pretrained layers as a possible confounder. To do so, we conduct a series of \textbf{localized reinitialization} experiments, where we take all blocks of three consecutive layers and either 1) \emph{reinitialize} those layers or 2) \emph{preserve} those layers while reinitializing the others in the network.\footnote{See the Appendix for more discussion and experiments where only one layer is reinitialized.}
These localized reinitializations help determine the extent to which BERT's different layers are either necessary (performance decreases when they are removed) or sufficient (performance is higher than random initialization when they are kept) for a specific level of performance. Again, BERT's residual connections permit the model to ignore reinitialized layers' outputs if they harm finetuning performance.

These results, shown in Figure \ref{fig:localized}, demonstrate that the earlier layers appear to be generally more helpful for finetuning relative to the later layers, even when controlling for the amount of finetuning data. However, there are strong task-specific effects: SST-2 appears to be particularly damaged by removing middle layers, while the effects on CoLA are distributed more uniformly. The effects on QNLI appear to be concentrated almost entirely in the first four layers of BERT---suggesting opportunities for future work on whether sparsity of this sort  indicates the presence of easy-to-extract features correlated with the task label. These results support the hypothesis that different kinds of feature processing learned during BERT pretraining are helpful for different finetuning tasks, and provide a new way to gauge similarity between different tasks.

\section{How vital is the ordering of pretrained layers?}
\label{sec:scrambles}

\begin{figure}
    \includegraphics[width=1\columnwidth]{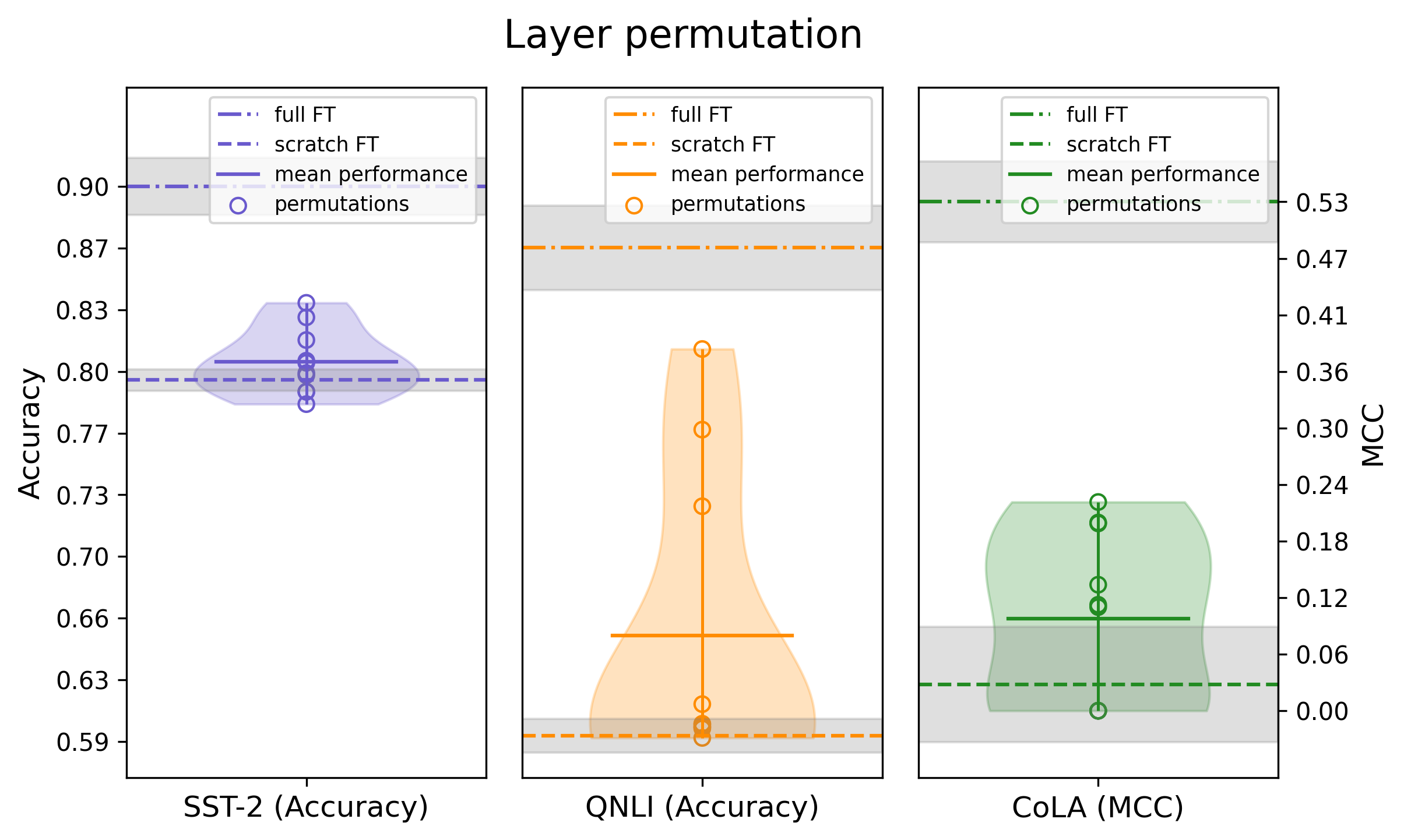}
    \caption{\textbf{Changing the order of pretrained layers harms finetuning performance significantly.} Dashed lines mark the performance of the original BERT model and the randomly-initialized model (surrounded by $\pm 2\sigma$ error bars). Circles denote finetuning performance for different layer permutations, while the solid line denotes the mean across runs (with 95\% CIs). The curved shaded region is a kernel density plot, which illustrates the distribution of outcomes. Finetuning trials with 5k examples.}
    \label{fig:scrambles}
\end{figure}

We also investigate whether the success of BERT depends mostly on learned \emph{inter-layer phenomena}, such as learned feature processing pipelines \citep{tenney2019bert}, or \emph{intra-layer phenomena}, such as a learned feature-agnostic initialization scheme which aid optimization \citep[e.g.][]{glorot2010understanding}. To approach this question, we perform several \textbf{layer permutation} experiments, where we randomly shuffle the order of BERT's layers before finetuning. The degree that finetuning performance is degraded in these runs indicates the extent to which BERT's finetuning success is dependent on a learned composition of feature processors, as opposed to providing better-initialized individual layers which would help optimization anywhere in the network.

These results, plotted in Figure \ref{fig:scrambles}, show that scrambling BERT's layers reduces their finetuning ability to not much above a randomly-initialized network, on average. This decrease suggests that BERT's transfer abilities are highly dependent on the intra-layer interactions learned during pretraining.

We also test for correlation of performance between tasks.  We do this by comparing task-pairs for each permutation, as we use the same permutation for the $n$th run of each task. The high correlation coefficients for most pairs shown in Table \ref{table:scrambles} suggest that BERT finetuning relies on similar inter-layer structures across tasks.

\begin{table}[h]
    \centering
    \begin{tabular}{lll}
        \toprule
        Tasks compared & Spearman & Pearson \\
        \midrule
           SST-2, QNLI & 0.72 (0.02) & 0.46 (0.18) \\
           SST-2, CoLA & 0.74 (0.02) & 0.77 (0.01) \\
            QNLI, CoLA & 0.83 (0.00) & 0.68 (0.03) \\
        \bottomrule
    \end{tabular}
    \caption{\textbf{Specific permutations of layers have similar impacts on finetuning across tasks.} Paired correlation coefficients between task performances for the same permutations. Two-sided $p$-value in parentheses (N=10).}
    \label{table:scrambles}
\end{table}

\section{Conclusion}
\label{sec:conclusion}

We present a set of experiments to better understand how the different pretrained layers in BERT influence its transfer learning ability. Our results reveal the unique importance of transferability of parameters to successful transfer learning, distinct from the transferability of fixed representations assessed by probing. We also disentangle important factors affecting the role of layers in transfer learning: task vs. quantity of finetuning data, number vs. location of pretrained layers, and presence vs. order of layers.

While probing continues to advance our understanding of linguistic structures in pretrained models, these results indicate that new techniques are needed to connect these findings to their potential impacts on finetuning. The insights and methods presented here are one contribution toward this goal, and we hope they enable more work on understanding why and how these models work.

\section{Acknowledgements}

We would like to thank Dan Jurafsky, Pranav Rajpurkar, Shyamal Buch, Isabel Papadimitriou, John Hewitt, Peng Qi, Kawin Ethayarajh, Nelson Liu, and Jesse Michel for useful discussions and comments on drafts. This work was supported in part by DARPA under agreement FA8650-19-C-7923.

\bibliography{emnlp2020}
\bibliographystyle{acl_natbib}

\clearpage

\appendix

\section{Code}

Our code is available at \url{https://github.com/dgiova/bert-lm-transferability}.

\section{Reinitialization}

We reinitialize all parameters in each layer, except those for layer normalization \citep{ba2016layer}, by sampling from a truncated normal distribution with $\mu = 0, \sigma = 0.02$ and truncation range $(-0.04, 0.04)$. For the layer norm parameters, we set $\beta=0, \gamma=1$. This matches how BERT was initialized (see the original BERT \href{https://github.com/google-research/bert/blob/master/modeling.py#L377}{code on GitHub} and the corresponding TensorFlow
\href{https://www.tensorflow.org/api_docs/python/tf/random/truncated_normal}{documentation}).

\section{Subsampling, number of trials, and error bars}
\begin{figure}[h]
    \includegraphics[width=\columnwidth]{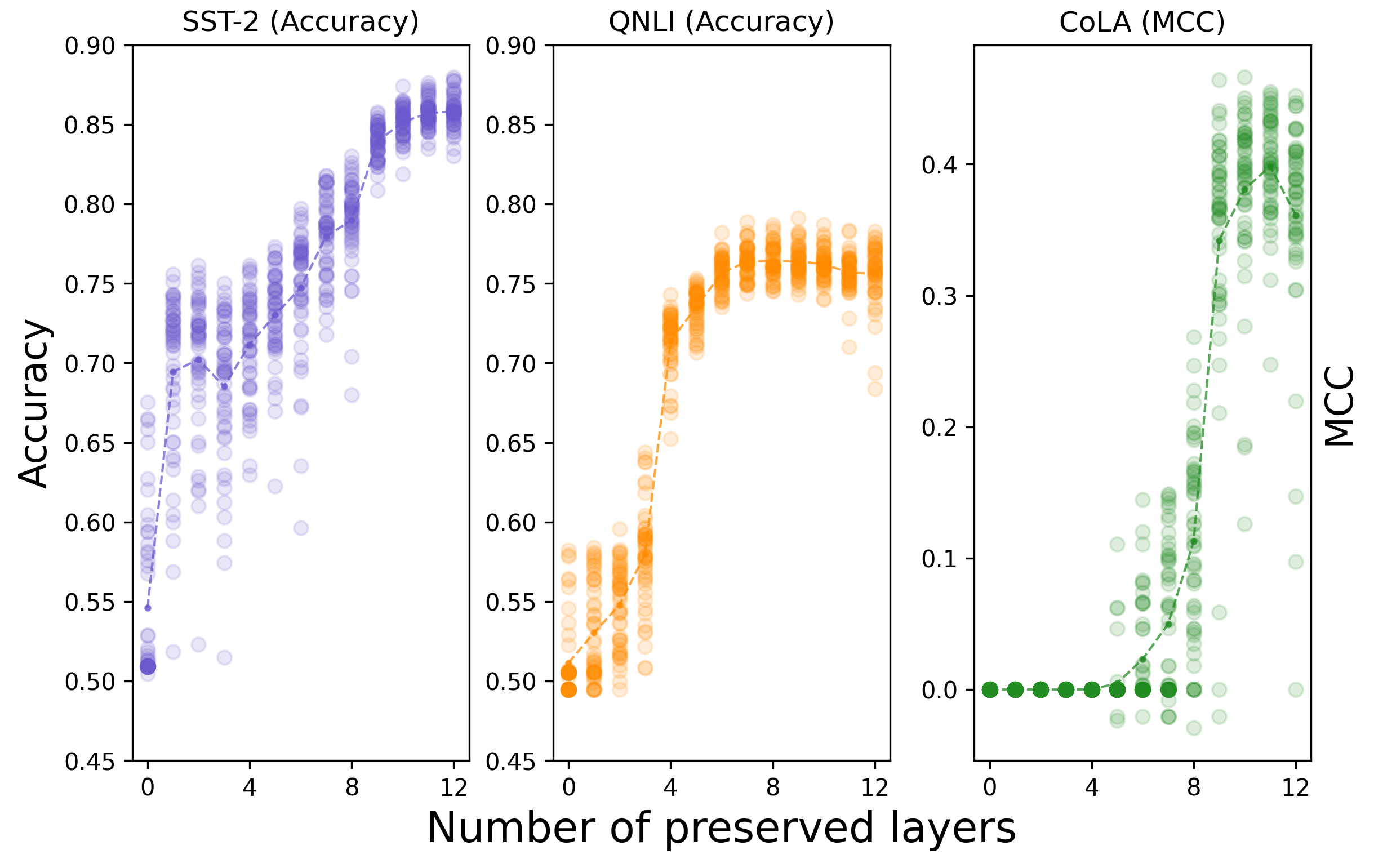}
    \caption{Finetuning results after reinitializing all layers past layer $k$ in BERT-Base. 12 shows the full BERT model, while 0 shows a model with all layers reinitialized. Scatterplot of 50 trials per layer shown for subsampled dataset size 500. Dotted line shows the mean.}
    \label{fig:scatter-500}
\end{figure}

The particular datapoints subsampled can have a large impact on downstream performance, especially when data is scarce. To capture the full range of outcomes due to subsampling, we randomly sample a different dataset for each trial index. Due to this larger variation when data is scarce, we perform 50 trials for the experiments with 500 examples, while we perform three trials for the other incremental reinitialization experiments. A scatterplot of the 500-example trials is shown in Figure \ref{fig:scatter-500}. For the localized reinitialization experiments, we perform ten trials each. 

Error bars shown on all graphs in the main text are 95\% confidence intervals calculated with a t-distribution. 

\section{Localized reinitializations of single layers}
\begin{figure}[h]
    \includegraphics[width=\columnwidth]{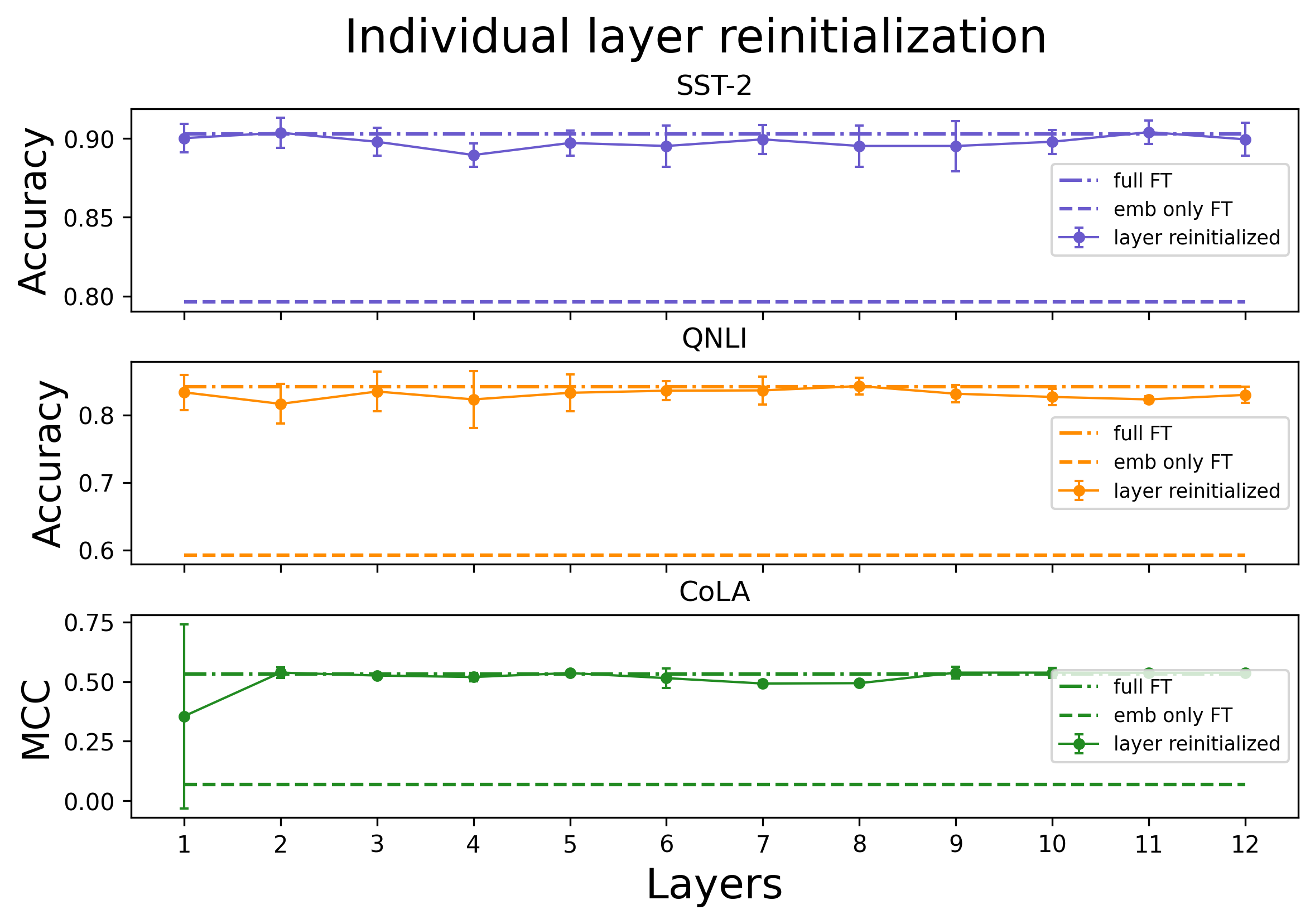}
    \caption{Performance on finetuning tasks after reinitializing an individual layer of BERT. Error bars are $\pm 2$ standard deviations.}
    \label{fig:localized-individual}
\end{figure}

We also experiment with performing our localized reinitialization experiments at the level of a single layer. To do so, we perform three trials of reinitializing each layer $k \in \{1 \ldots 12\}$ and then finetuning on each of the three GLUE tasks. Our results are plotted in Figure \ref{fig:localized-individual}. Interestingly, we observe little effect on finetuning performance from reinitializing each layer (except for reinitializing the first layer on CoLA performance). This lack of effect suggests either redundant information between layers or that the ``interface'' exposed by the two neighboring layers somehow beneficially constrains optimization.

\section{Number of finetuning epochs}
\citet{he2019rethinking} found that much or all of the performance gap between an ImageNet-pretrained model and a model trained from random initialization could be closed when the latter model was trained for longer. To evaluate this, we track validation losses up to ten epochs in our incremental experiments, for $k \in \{0, 6, 12\}$ across all tasks and for 500 and 5k examples. We find minimal effects of training longer than three epochs for the subsamples of 5k, but find improvements of several percentage points for training for five epochs for the trials with 500 examples. Thus, for the trials of 500 in Figure \ref{fig:progressive}, we train for five epochs, while training for three epochs for all other trials. We train our probing experiments (8 trials per layer) with early stopping for a maximum of 40 epochs on the full dataset.

\section{Higher learning rate for reinitialized layers}

\begin{figure}[h]
    \includegraphics[width=\columnwidth]{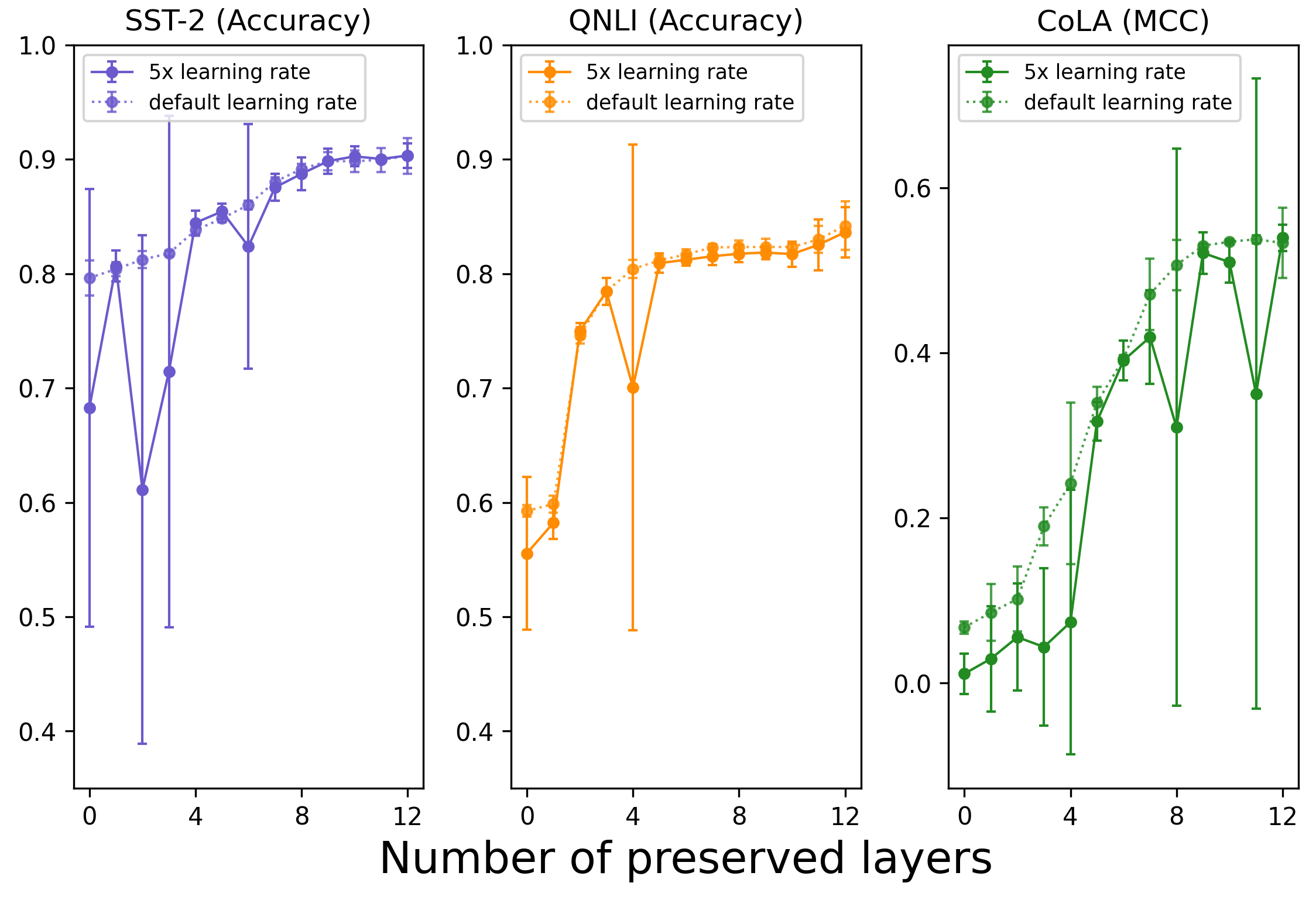}
    \caption{Finetuning the reinitialized layers with a larger learning rate does not improve finetuning performance. Error bars are $\pm 2$ standard deviations.}
    \label{fig:lr-5x}
\end{figure}

In their reinitialization experiments on a convolutional neural network for medical images, \citet{raghu2019transfusion} found that a 5x larger rate on the reinitialized layers enabled their model to achieve higher finetuning accuracy. To evaluate this possibility in our setting, we increase the learning rate by a factor of five for the reinitialized layers. The results for our incremental reinitializations are plotted in Figure \ref{fig:lr-5x}. A higher learning rate appears to increase the variance of the evaluation metrics while not improving performance. Thus, we keep the learning rate the same across layers. 

\section{Layer norm}
Because the residual connections around each sublayer in BERT are of the form $\mathrm{LayerNorm}(x + \mathrm{Sublayer}(x))$, reinitializing a particular layer neutralizes the effect of the last layer norm application from the previous layer in a way that cannot be circumvented through the residual connections. However, for brevity we simply refer to ``reinitializing a layer'' in this paper.

\begin{figure}
    \includegraphics[width=\columnwidth]{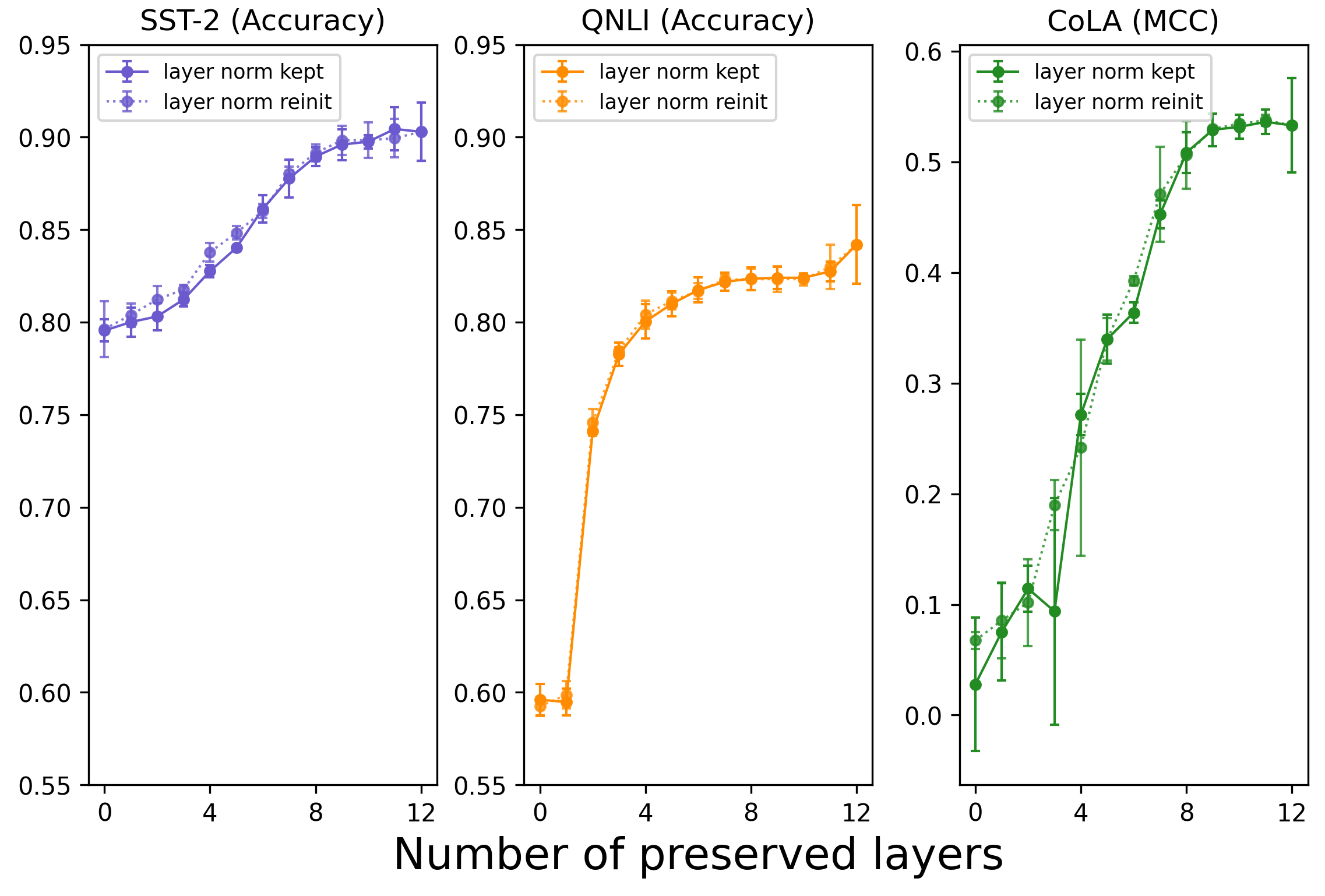}
    \caption{Preserving the layer norm parameters when reinitializing each layer does not improve finetuning performance. Error bars are $\pm 2$ standard deviations.}
    \label{fig:progressive-keep-ln}
\end{figure}

We also assessed whether preserving the layer norm parameters in each layer might aid optimization. To do so, we preserved these parameters in our incremental trials with 5k examples. These trials are plotted in Figure \ref{fig:progressive-keep-ln}, and demonstrate that preserving layer norm does not aid (and may even harm) finetuning of reinitialized layers.

\section{Dataset descriptions and statistics}

We display more information about the finetuning datasets, including the full size of the datasets, in Table \ref{tab:data}.  

\begin{table*}
    \centering
    \caption{\textbf{Task description and statistics.} SST-2 and CoLA are single sentence classification tasks, while QNLI is a sentence-pair classification task.}
    \begin{tabular}{p{1cm}|p{1cm}|p{1cm}|p{7cm}|p{2cm}} \toprule
         Task & \#~Train & \# Val & Input, labels & Eval metric \\
         \midrule
         SST-2 & $67$k & $872$k & sentence, \{positive, negative\} & Accuracy \\
         QNLI & $105$k & $5.4$k & (question, paragraph), \{answer, non-answer\} & Accuracy \\
         CoLA & $8.5$k & $1$k & sentence, \{acceptable, not acceptable\} & MCC\\
         \bottomrule 
    \end{tabular}
    \label{tab:data}
    \end{table*}

\section{Additional experimental information}

\subsection{Link to data}
Scripts to download the GLUE data can be found at \url{https://github.com/nyu-mll/jiant/blob/master/scripts/download_glue_data.py}.

\subsection{Computing infrastructure}
All experiments were run on single Titan XP GPUs.

\subsection{Model}

We use the BERT-Base uncased model (110 million parameters) from \url{https://huggingface.co/transformers/pretrained_models.html}.

\subsection{Average runtime}
Average runtime for each approach:
    \begin{enumerate}
        \item \textbf{500 incremental}: 0.3 min / epoch * 5 epochs / trial * 50 trials / layer * 12 layers / task * 3 tasks $\approx$ 45 GPU-hrs
        \item \textbf{5k incremental}: 3 min / epoch * 3 epochs / trial * 3 trials / layer * 12 layers / task * 3 tasks $\approx$ 16 GPU-hrs.
        \item \textbf{50k incremental}: 30 min / epoch * 3 epochs / trial * 3 trials / layer * 12 layers / task * 3 tasks $\approx$ 7 GPU-days.
        \item \textbf{5k localized (block size 3)}: 3 min / epoch * 3 epochs / trial * 3 trials / layer * 10 layers / task * 3 tasks $\approx$ 14 GPU-hrs
        \item \textbf{Probing}: 2.8 min / epoch * 40 epochs / trial * 8 trials / layer  * 12 layers / task * 3 tasks $\approx$ 22 GPU-days. \emph{Note: 2.8 min / epoch is an average across layers and tasks. Earlier layers take less time than later ones because layers after the target layer do not need to be computed.}
    \end{enumerate}

\subsection{Evaluation method}
To evaluate the performance of our method, we compute \emph{accuracy} for SST-2 and QNLI and \emph{Matthews Correlation Coefficient} \citep{matthews1975comparison} for CoLA. We compute these metrics always on the official validation sets, which are never seen by the model during training.

Accuracy measures the ratio of correctly predicted labels over the size of the test set. Formally:

$\mathrm{accuracy} = \frac{TP + TN}{TP + TN + FP + FN}$

Since CoLA presents class imbalances, MCC is used, which is better suited for unbalanced binary classifiers \citep{warstadt2019neural}. It measures the correlation of two Boolean distributions, giving a value between -1 and 1. A value of 0 means that the two distributions are uncorrelated, regardless of any class imbalance.
$MCC = \frac{(TP \cdot TN) - (FP \cdot FN))}{\sqrt{(TP + FP)(TP + FN)(FP + TN)(TN + FN)}}$

\subsection{Hyperparameters}
We performed one experiment with a 5x learning rate and implemented early stopping to choose the number of epochs for the probing experiments.

For batch size and learning rate, we kept the default parameters for all tasks:
\begin{itemize}
\item Learning rate: 2e-5 
\item Batch size: 8
\end{itemize}

\end{document}